\documentclass{article}

\usepackage{arxiv}

\usepackage[utf8]{inputenc} 
\usepackage[T1]{fontenc}    
\usepackage{hyperref}       
\usepackage{url}            
\usepackage{booktabs}       
\usepackage{amsfonts}       
\usepackage{nicefrac}       
\usepackage{microtype}      
\usepackage{lipsum}		
\usepackage{graphicx}
\usepackage{natbib}
\usepackage{doi}
\usepackage{multirow}
\usepackage{amsmath}
\usepackage{tabularx}
\usepackage{makecell}
\usepackage{adjustbox}

\usepackage{caption} 
\title{All in one timestep: Enhancing Sparsity and Energy efficiency in Multi-level Spiking Neural Networks}

\date{} 					

\author{ {Andrea Castagnetti} \\
	Université Côte d'Azur\\
	LEAT\\
	Sophia Antipolis, France \\
	\texttt{andrea.castagnetti@univ-cotedazur.fr} \\
	\And
	{Alain Pegatoquet} \\
	Université Côte d'Azur\\
	LEAT\\
	Sophia Antipolis, France \\
	\texttt{alain.pegatoquet@univ-cotedazur.fr} \\
	\And
        {Benoît Miramond} \\
	Université Côte d'Azur\\
	LEAT\\
	Sophia Antipolis, France \\
	\texttt{benoit.miramond@univ-cotedazur.fr} \\
}

\hypersetup{
pdftitle={All in one timestep: Enhancing Sparsity and Energy efficiency in Multi-level Spiking Neural Networks},
pdfauthor={Andrea Castagnetti, Alain Pegatoquet, Benoît Miramond},
pdfkeywords={Spiking Neural Networks, multi-level spikes, Spiking ResNets, low latency, sparsity, low-power artificial intelligence},
}

\begin{document}
\maketitle

\begin{abstract}
Spiking Neural Networks (SNNs) are one of the most promising bio-inspired neural networks models and have drawn increasing attention in recent years. The event-driven communication mechanism of SNNs allows for sparse and theoretically low-power operations on dedicated neuromorphic hardware. However, the binary nature of instantaneous spikes also leads to considerable information loss in SNNs, resulting in accuracy degradation.
To address this issue, we propose a multi-level spiking neuron model able to provide both low-quantization error and minimal inference latency while approaching the performance of full precision Artificial Neural Networks (ANNs). Experimental results with popular network architectures and datasets, show that multi-level spiking neurons provide better information compression, allowing therefore a reduction in latency without performance loss. When compared to binary SNNs on image classification scenarios, multi-level SNNs indeed allow reducing by 2 to 3 times the energy consumption depending on the number of quantization intervals. On neuromorphic data, our approach allows us to drastically reduce the inference latency to 1 timestep, which corresponds to a compression factor of 10 compared to previously published results.
At the architectural level, we propose a new residual architecture that we call Sparse-ResNet. Through a careful analysis of the spikes propagation in residual connections we highlight a spike avalanche effect, that affects most spiking residual architectures. Using our Sparse-ResNet architecture, we can provide state-of-the-art accuracy results in image classification while reducing by more than 20\% the network activity compared to the previous spiking ResNets.
\end{abstract}


\section{Introduction}
Spiking Neural Networks (SNNs) are considered as the the third generation of Artificial Neural Networks (ANNs). SNNs focus on the encoding and the processing of the information using binary and asynchronous signals known as spikes. This computing paradigm, inspired by the messaging mechanism used by biological neurons, is thought to be amongst the sources of the energy efficiency of the brain. However, the binary nature of spikes also leads to considerable information loss, i.e. quantization errors, causing performance degradation compared to ANNs using floating-point operations. SNNs quantization error can be reduced by increasing the number of timesteps, therefore the latency over the network. However, with a longer conversion time, more spikes are generated, thus increasing the energy consumption as well. Several techniques have been proposed to minimize both the quantization error and the latency of SNNs. These approaches can be either applied to the ANN-to-SNN conversion or directly during the SNN training using the surrogate gradient (SG) method. In \cite{li_quantization_2022} and \cite{rathi_diet-snn_2021} the authors adopt an ANN-to-SNN conversion scheme where the firing threshold
of the spiking neurons is optimized after conversion to better match the distribution of the membrane potential. In \cite{castagnetti_trainable_2023} a SNN is trained using SG and the Adaptive Integrate-and-Fire (ATIF) neuron. This ATIF neuron has been proposed as an alternative to the original Integrate and Fire (IF) neuron, the firing threshold ($V_{th}$) being a learnable parameter rather than an empirical hyper-parameter. In \cite{guo_recdis-snn_2022}, the authors also use SG to train the SNN, but introduce a distribution loss to shift the membrane potential distribution into the conversion range of the spiking neurons. With these approaches, requiring only few timesteps, it is
possible to get SNNs with almost no accuracy loss when
compared to the equivalent ANNs. As an example, a SNN trained on 4 timesteps can achieve an accuracy only 1\% below the equivalent non-quantized, i.e. Floating Point (FP), ANNs on CIFAR-10 \cite{li_quantization_2022}. To further decrease the latency, recent approaches propose to go beyond binary spikes and introduce multi-level spiking neurons. This mechanism expands the output of spiking neurons from a single bit to multiple bits, thus increasing the information that can be transmitted at each timestep. In \cite{xiao_multi-bit_2024} for instance, it has been shown that using a 4-level spiking neuron and one timestep, the same level of accuracy of a well optimized binary SNNs trained on 4 timesteps \cite{li_quantization_2022} can be achieved. Most of the previous works only focus on the SNN latency. However, it has been shown \cite{lemaire_analytical_2023, dampfhoffer_are_2023} that besides the latency, another important parameter that has to be optimized to improve the energy efficiency is the sparsity of the network, in other words the number of spikes, either binary or multi-level, generated during the inference. In this work we propose to enhance the sparsity of SNN from two points of view. First at the neuronal level, where we introduce a multi-level spiking neuron model that can seamlessly deal with both time and the spike value to reduce the quantization error. Then at the architectural level, where a novel spiking ResNet architecture is proposed. Through a careful analysis of the spike propagation in the residual layers of the architecture we highlight an effect that we call \emph{spike avalanche}. Events coming from the direct and residual paths sum up and create peaks of activity that can potentially decrease the sparsity of the SNN. Finally we compare binary and multi-level SNNs from the energy-efficiency point of view. Our analysis based on the metric proposed in \cite{lemaire_analytical_2023}, is intended to be independent from low-level implementation choices. Moreover, our energy estimation takes into account not only the synaptic operations but also all the memory accesses that take place in an event-driven execution scenario on a neuromorphic hardware accelerator. We therefore provide energy estimation results that are closer to what could be reasonably obtained on a real hardware implementation. This approach is essential to avoid overestimating the energy gains. 

The main contributions of our work are listed below:
\begin{itemize}
    \item We propose a multi-level model of an IF spiking neuron compatible with SNN direct training using SG.
    \item We train SNNs on different image classification problems and characterize the corresponding spiking activity. We provide state-of-the-art accuracy results on CIFAR-10/100 image datasets, using only 1 timestep while reducing the energy consumption by a factor of 3 compared to an equivalent ANN.
    \item On neuromorphic data we provide a new state of the art latency/accuracy results for the CIFAR-10-DVS dataset. Here we approach the previously published best accuracies, which are obtained with 10 or more timesteps, while compressing the latency to 1 timestep. 
    \item We introduce Sparse-ResNet, a new spiking residual architecture. Our experimental results show that Sparse-ResNet can achieve accuracy equivalent to the state of the art SEW-ResNet \cite{fang_deep_2021} while reducing the spiking activity by more than 20\% on the CIFAR-10 dataset.
\end{itemize}

\section{Related Works}
\subsection{SNN training}
Training large SNNs models on modern complex datasets has been the subject of an increasing number of studies in the recent years. Two methods are typically used to obtain an efficient SNN: ANN-SNN conversion or direct training with surrogate gradients.
The ANN-SNN conversion method is based on the idea that firing rates of spiking neurons should match the activations of analog, i.e. Floating Points, neurons.
Earlier works proposed to directly convert from FP ANN models to SNNs. \cite{diehl_fast-classifying_2015} uses weight normalization and spiking neuron threshold balancing to minimize the conversion loss. In \cite{rueckauer_conversion_2017} and \cite{han_rmp-snn_2020} the authors propose to adopt reset-by-subtraction, i.e. soft-reset, spiking neurons to lower the conversion loss. Moreover they show that the SNN accuracy can be improved by using real inputs instead of Poisson spike trains. In \cite{stanojevic_exact_2023} the ANN is converted in an equivalent SNN that uses a Time-To-First-Spike (TTFS) coding method to propagate the information between the spiking neurons. However, directly converting a FP ANN to an SNN typically leads to 
a very high latency solution. Nevertheless, it has been recently shown in \cite{wang_universal_2024} that it is possible to achieve high accuracy at low latency using a specific neuron initialization, called data-based neuronal initialization (DNI) and by taking into account the asynchronous transmission characteristics of SNNs.

Another line of works has shown that to achieve very low latency and high accuracy at the same time, the quantization process has to be taken into account directly when training the ANN model. In \cite{li_quantization_2022} the ANN model is first trained using a Quantization-Aware-Training (QAT) procedure, and then converted into an equivalent SNN. Similarly \cite{ding_optimal_2021} replaces the ReLU activation functions with the proposed quantization clip-floor-shift (QCFS) activation. The ANN is trained using stochastic gradient descent and a straight-through estimator \cite{bengio_estimating_2013} is used to approximate the derivative of the quantized activation function. In \cite{yang_cs-qcfs_2025} the conversion method proposed in \cite{ding_optimal_2021} is extended to Channel-wise quantization. 

Instead of converting from a pre-trained ANN, the direct training method \cite{neftci_surrogate_2019} proposes to directly train the SNN using standard back-propagation.
A surrogate function is used, during gradient back-propagation, to replace the binary non-linearity of spiking neurons. This allows gradient flowing thus making back-propagation possible in the spiking domain. Intuitively, direct training of SNNs with surrrogate gradient share some similarities with the quantized ANN to SNN conversion methods discussed above. The spike information compression is analyzed in \cite{castagnetti_trainable_2023, castagnetti_spiden_2023, li_quantization_2022}, where the quantization scheme is described as a function of the neuron parameters, i.e. $V_{th}$ and the number of timesteps, i.e. $T$. 
A spiking neuron can actually be considered as a uniform quantizer that discretizes its input into $T+1$ quantization intervals. The direct training method leverages this fact and thus optimizes the quantization mapping while seeking to minimize the task loss. Several works improve the accuracy/latency trade-off by using techniques to relieve the gradient exploding/vanishing problem incurred by the back-propagation-through-time (BPTT) algorithm. \cite{guo_recdis-snn_2022} introduces a membrane potential distribution loss to  penalize undesired membrane potential shift thus alleviating the gradient vanishing or explosion. \cite{li_directly_2024} proposes Masked Surrogate Gradient (MSG) to keep the balance between the gradient mismatch, caused by a wide surrogate function, and the vanishing gradient risks incurred by a narrow surrogate function.

\subsection{Quantization noise in spiking neurons}
Several other works try to improve the accuracy/latency trade-off of SNNs by minimizing the quantization noise introduced by the spiking neurons. \cite{guo_im-loss_2022} introduces an Information-Maximization loss to minimize the information-loss caused by the quantization from the full-precision tensor, i.e. the input of the spiking neurons, to the spike tensor. \cite{castagnetti_trainable_2023} proposes to minimize the quantization error by learning the firing threshold ($V_{th}$) of the spiking neuron during training. In \cite{chen_high-performance_2024} a time-based coding scheme called At-most-two-spikes Exponential Coding (AEC) is introduced. AEC neurons employ quantization-compensating spikes to improve coding accuracy and capacity. Finally, \cite{chen_high-performance_2024} proposes a neuron model that exploits the option of temporal coding with spike patterns, where the timing of a spike transmits extra information.

\subsection{Multi-level spiking neurons}
To further decrease the latency, recent approaches propose to go beyond binary spikes and introduce multi-level spiking neurons. This mechanism expands the output of spiking neurons from a single bit to multiple bits, thus increasing the information that can be transmitted at each timestep.
In \cite{guo_ternary_2023} the authors propose a ternary spiking neuron that transmits information with \{-1, 0, 1\} spikes. A similar ternary coding is proposed in \cite{sun_deep_2022} but only with positive values. Moreover, in multi-level spiking neurons the spike is extended to a fixed-point unsigned binary number \cite{feng_multi-level_2022, xiao_multi-bit_2024}. In \cite{xiao_multi-bit_2024}, it has been shown that using a 4-level spiking neuron and one timestep, the same level of accuracy of a well optimized binary SNNs trained on 4 timesteps \cite{li_quantization_2022} can be achieved. \cite{wang_mt-snn_2024} introduces a spiking neuron model with multiple thresholds, called MT-SNN that can generate multiple spike sequences. MT-SNN can achieve a higher accuracy with fewer timesteps since the precision loss caused by the shorter latency, i.e. $T$, is restored by the information encoding obtained using multiple thresholds.

\subsection{Event-based processing, sparsity and energy consumption estimation}
SNNs hold the promise of lower energy consumption in embedded hardware due to their sparse event-driven spike-based computations compared to traditional ANNs. In SNNs the neuron computation consists in accumulating the spikes, weighted by the synaptic connections into the membrane potential and firing an output spikes when the membrane potential exceeds $V_{th}$. In its simplest form the spiking neuron operations only require accumulate (AC) operations while artificial neurons in ANNs perform multiply-and-accumulate (MAC) operations between input activations and weights. Most research works, considering either binary \cite{wang_universal_2024, kim_spiking-yolo_2020, rathi_diet-snn_2021} or multi-level SNN \cite{guo_ternary_2023, xiao_multi-bit_2024, wang_mt-snn_2024}, posits that the energy efficiency of SNNs is associated with the energy savings obtained by replacing MACs with ACCs for synaptic operations coupled with the inherent sparsity of the SNN models. However, this point of view has been questioned in recent works \cite{lemaire_analytical_2023, dampfhoffer_are_2023} where it has been shown that the energy cost of synaptic operations is negligible compared to the one of transferring data, i.e. synaptic weights and neuron potentials, from or to the memory. As an example, \cite{dampfhoffer_are_2023} estimates that for an SNN with a VGG16 topology the energy consumed by the synaptic operations represents only 1\% of the total energy consumption. Moreover, these works point out that the main advantage of SNNs compared to ANNs (on digital hardware) comes primarily from exploiting the sparsity of spikes and not from the replacement of MAC by ACC operations \cite{dampfhoffer_are_2023}. Based on these results we believe that to improve the SNNs energy/accuracy trade-off, the classical latency/accuracy trade-off has to be revisited. We then propose to analyze the SNNs behavior by focusing on the sparsity as well as its native ability to process data in the time domain. In the following we first analyze how information is coded in binary spiking neuron. Then we introduce a multi-level spiking neuron model compatible with direct training using SG. We characterize the quantization noise of the proposed neuron model. We then move to network level and analyze the propagation of spikes in residual layers. We show that without a careful handling of residual connections in SNNs, the sparsity of residual networks can be compromised, thus increasing the energy consumption.

\section{Preliminaries} \label{sec:preliminaries}
In this section, we introduce the Integrate-and-Fire (IF) binary spiking model and describe the encoding and decoding process used in SNNs.
\subsection{Information coding with spiking neurons}
An IF spiking neuron implements the ReLU non-linearity ($max(0,x)$) and discretizes its input signal into spikes. The following equations describe the Integrate-and-Fire (IF) neuron model with soft-reset:
\begin{align}
    H^{l}(t) &= V^{l}(t-1) + i^{l}(t) \label{eq:IF_H}\\
    i^{l}(t) &= z^{l-1}(t)\,W^{l} + b^{l} \label{eq:IF_i}\\
    z^{l}(t) &= \Theta(H^{l}(t) - V_{th}) \label{eq:IF_z}\\
    V^{l}(t) &= H^{l}(t)\,(1 - z^{l}(t)) + (H^{l}(t) - V_{th})\,z^{l}(t) \label{eq:IF_V}
\end{align}

The previous equations describe the behavior of the IF neuron at the output of the layer $l$ of the SNN. The membrane potential after the input integration and after the reset operation are represented by $H^{l}(t)$ and $V^{l}(t)$ respectively. The binary spiking output at time $t$ is represented by $z^{l}(t)$.
The input of the spiking neuron, $i^{l}(t)$, is the sum of the weighted spikes coming from the layer $l-1$ plus a constant bias. As shown in Eq. \ref{eq:IF_H} and \ref{eq:IF_i}, the synaptic operations are implemented with ACC in a spiking neuron. Eq. \ref{eq:IF_z} and \ref{eq:IF_V} describe the generation of a spike and the soft-reset operation respectively. The function $\Theta(\cdot)$ represents the Heaviside step function used in the forward pass. The surrogate of $\Theta(\cdot)$, that is $\Theta^{\prime}(x) = \sigma^{\prime}(x)$, is used during the backward pass. In this work we use the sigmoid $\sigma(x, \alpha) = \frac{1}{1 - e^{-\alpha x}}$ as surrogate function. Where $x$ represents the membrane voltage of the spiking neuron and $\alpha$ is an hyper-parameter that modulates the width of the surrogate derivative.

The integrate-and-fire operations described by the previous equations, are repeated through $T$ timesteps. The output $y_s^{l}$ is then rate-decoded as follows:
\begin{equation}
    y_s^{l} = \frac{1}{T}\sum_{t=1}^T z^{l}(t)
    \label{eq:SNN_decoding}
\end{equation}
The Eq. \ref{eq:SNN_decoding} represents the firing-rate of the neuron. In this scheme the information is coded by the number of spikes generated by a spiking neuron over a fixed duration of time $T$. Moreover, since $z^{l}(t)$ is a binary variable, the decoded output $y_s^{l}$ is quantized into $T+1$ different values. The number of quantization intervals determines the distortion incurred during the quantization process. For a binary spiking neuron the quantization noise can be reduced by increasing $T$, i.e. the latency of the SNN. However, augmenting the latency leads to an increase of the SNN energy consumption as more spikes are generated. There is therefore a trade-off between the amount of quantization noise and the energy efficiency of the SNN. Our objective is therefore to reduce the quantization noise without increasing $T$. The next section proposes a solution for that problem by introducing the multi-level spiking neuron model.

\section{Methods}
\subsection{Multi-level spiking neuron model} \label{sec:multi-level-neuron}
\begin{figure*}[htbp]
\centerline{\includegraphics[width=.95\textwidth]{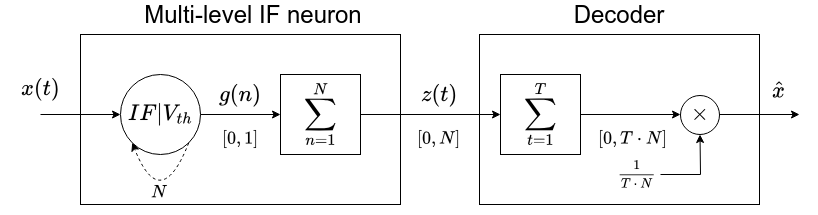}}
\caption{Multi-level IF neuron model and the associated decoding scheme.}
\label{fig:multi_level_neuron}
\end{figure*}

The neuron model and the associated decoding scheme are shown in Fig. \ref{fig:multi_level_neuron}.
This model is characterized by two parameters: the firing threshold $V_{th}$ and the number of levels of a spike, that we note $N$. The multi-level spiking neuron is composed by a standard IF neuron, described by the Eq. \ref{eq:IF_H}-\ref{eq:IF_V} given in Sec. \ref{sec:preliminaries}.
The spike generation process is divided into two phases. At each timestep the IF neuron is iteratively charged, $N$ times, with the current input $x(t)$. Through this operation, the current timestep is indeed subdivided into $N$ intervals, that we call \emph{micro-timesteps}. Then, the internal membrane potential of the IF neuron is discharged through the $N$ micro-timesteps to generate $g(n)$ binary spikes. The value of the spike emitted at the end of the timestep is computed by summing up all the binary spikes $g(n)$ generated through the discharge process:
\begin{equation}
    z(t) = \sum_{n=1}^{N} g(n)
\end{equation}
It is worth noticing that only the valued spike $z(t)$ is communicated to the next layer. The binary spikes $g(n)$ are internal to the spiking neuron and do not contribute to the overall activity of the network.

\subsection{Quantization error analysis} \label{sec:quantization}
\begin{figure}[htbp]
\centerline{\includegraphics[width=.5\textwidth]{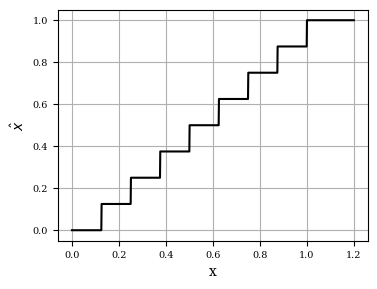}}
\caption{Quantization function of a multi-level IF with soft-reset ($T = 2$, $N = 4$, $V_{th} = 1.0$). We can observe that the output of the neuron has exactly $N \times T + 1$ quantization levels and the output saturates when the input equals $V_{th}$.}
\label{fig:multi_level_quantization}
\end{figure}

The spiking neuron operation described in Sec. \ref{sec:multi-level-neuron} can be repeated through an arbitrarily number of T timesteps, where the output at each timesteps is a valued spike $z(t) \in [0, N]$. Moreover, the model can be reverted to an IF binary spiking neuron by setting $N=1$. In that case, the discharge process is executed only once and the output at each timesteps is a binary spike $z(t) \in [0, 1]$. Finally, the output of the spiking neuron can be decoded as follows:
\begin{equation}
    \hat{x} = \frac{1}{N\,T}\,\sum_{t=1}^{T}z(t)
\end{equation}
Overall, the conversion process leads to a discretization of the input $x$. 
When the input of the neuron is a constant signal, that is $x(t) = x$, we can observe that the quantization function of the multi-level spiking neuron, shown in Fig. \ref{fig:multi_level_quantization}, has exactly $N \times T + 1$ quantization levels. For the binary spiking neuron, that is when $N=1$, there are instead $T + 1$ uniform quantization intervals \cite{castagnetti_trainable_2023, li_quantization_2022}.
A multi-level spiking neuron can therefore communicate at each timestep $\log_2(N)$ bits while the binary spiking neuron is limited to only 1 bit of information. With multi-level spiking neurons it is thus possible to decrease the quantization error without increasing the latency of the SNN, thus achieving a better information compression without hindering, in theory, the computational efficiency of the SNN.

\subsection{Spikes propagation in residual connections} \label{sec:spikes_propagation_qualitative}
\begin{figure*}[htbp]
\centerline{\includegraphics[width=.9\textwidth]{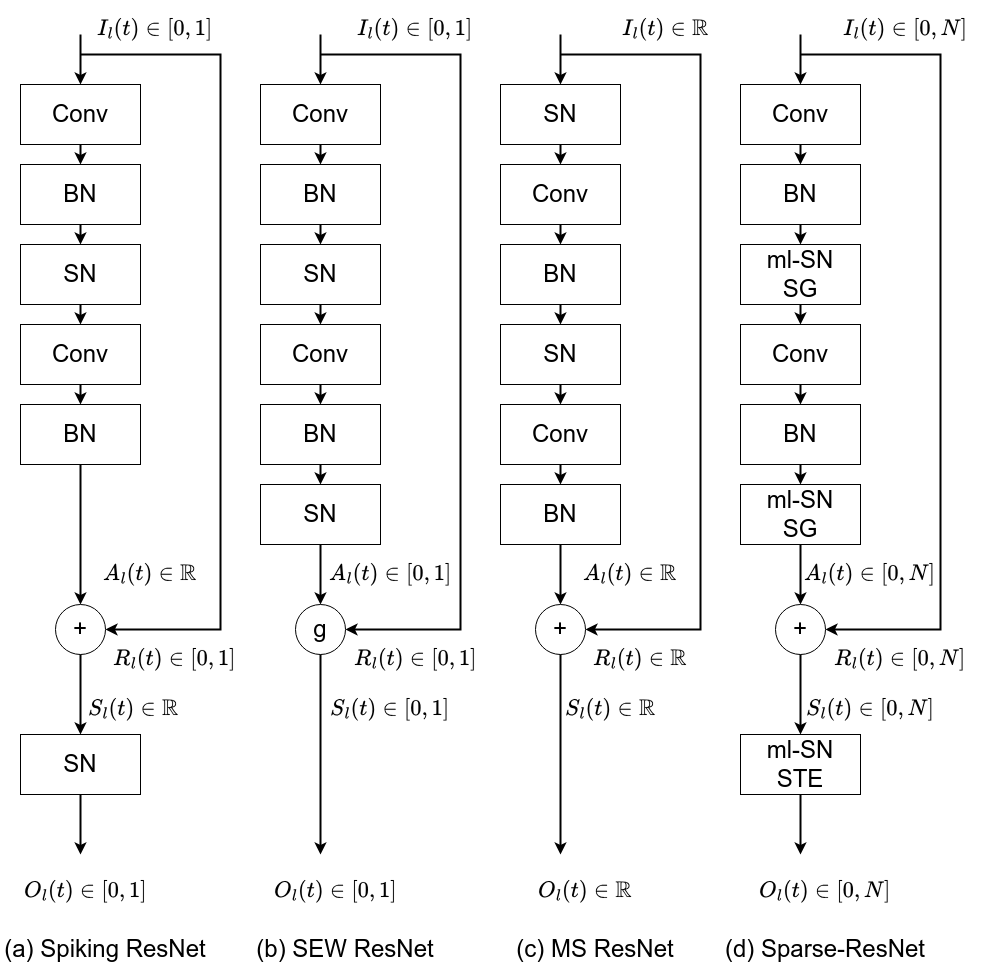}}
\caption{Residual blocks in (a) Spiking ResNets \cite{hu_spiking_2023}, (b) SEW-ResNets \cite{fang_deep_2021}, (c) MS-ResNets \cite{hu_advancing_2023} and the proposed Sparse-ResNets. \emph{SN} stands for binary spiking neuron while \emph{ml-SN/STE} and \emph{ml-SN/SG} means a multi-level spiking neuron with a Straight-Through Estimator and a surrogate backward function respectively.}
\label{fig:spiking_resnets}
\end{figure*}

\begin{figure}[htbp]
\centerline{\includegraphics[width=.5\textwidth]{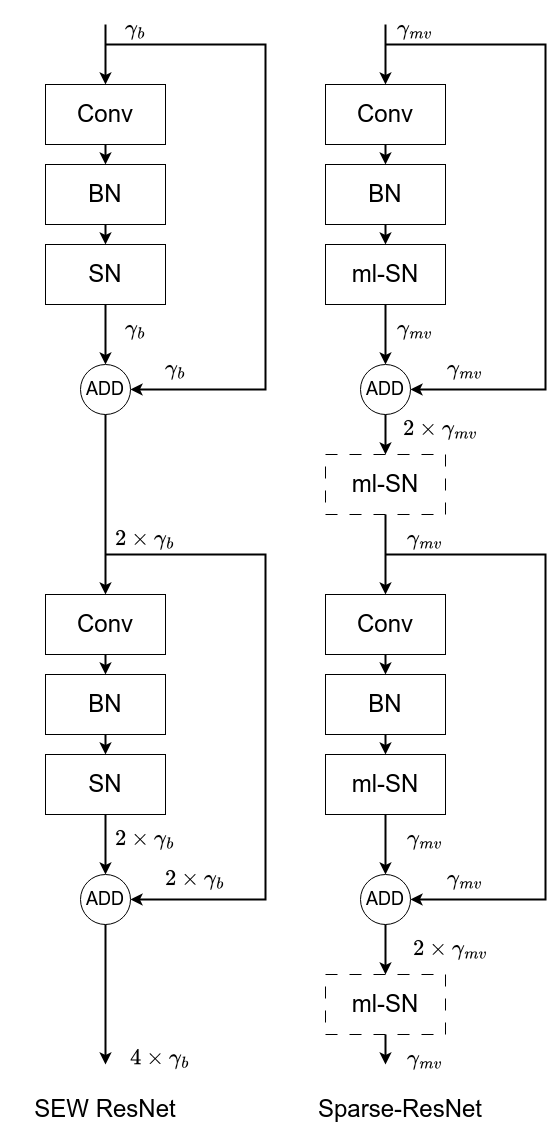}}
\caption{Spike propagation in residual connections. Here $\gamma_b$ and $\gamma_{mv}$ represent the amount of binary and multi-level spikes at the input of the first residual connection of the network. Only one Convolutional block is represented in the direct path. In the SEW-ResNets \cite{fang_deep_2021}, the flow of spikes is added at the summation points, thus creating an exponential increase of the number of spikes that have to be processed by deeper layers. Using the barrier neuron (dotted in the figure), the Sparse-ResNets can limit the propagation of spikes without hindering the representation capacity of the network.}
\label{fig:avalanche_resnets}
\end{figure}

Residual learning and shortcuts connections have been evidenced as an important way of deepening ANNs architectures \cite{he_deep_2016} and are now widely adopted by almost any state-of-the-art neural network \cite{sandler_mobilenetv2_2019, tan_efficientnet_2020, dosovitskiy_image_2021}.
Particularly, ResNets architectures \cite{he_deep_2016} have been converted into SNNs in several previous works \cite{hu_spiking_2023, fang_deep_2021, hu_advancing_2023}. 
Conversion issues from ANNs have been studied in \cite{hu_spiking_2023}, while \cite{fang_deep_2021} focused on the difficulty of implementing identity mapping with spiking neurons. Finally in \cite{hu_advancing_2023} the authors explored the vanishing/exploding gradient problems that arise when using the direct training method.
The main residual architectures for SNNs are shown in Fig. \ref{fig:spiking_resnets}. 
In the original Spiking ResNet proposed in \cite{hu_spiking_2023} the only architectural modification consists in replacing the ReLU non-linearities with spiking neurons. Spiking ResNet converted from ANN achieves state-of-the-art accuracy on nearly all datasets, albeit with high latencies \cite{han_rmp-snn_2020}. On the other hand, several challenges have to be addressed to achieve the same accuracy levels with Spiking ResNets trained directly in the spike domain. MS-ResNet \cite{hu_advancing_2023} addresses this problem by implementing the residual connections between the membrane potentials of the spiking neurons. However, by directly connecting the membrane potential of the spiking neurons, the MS-ResNets cannot be considered a fully event-based network. On the other hand, SEW-ResNets \cite{fang_deep_2021} proposes to aggregate the spikes at the summation point of the residual connection. Using this topology, SEW-ResNets can indeed achieve comparable results with standard ResNets while maintaining the event-based communication scheme. However, as the spikes are sequentially transferred from one layer to the next, the spikes that come from the residual connection are added to the flow of spikes coming from the direct path. This effect, that we call \emph{spike avalanche} is illustrated in Fig. \ref{fig:avalanche_resnets}. 
In this simplified example, we consider that the number of input binary spikes in SEW ResNets, that we call $\gamma_b$, is not modified when processed by a convolutional layer. As shown, the $\gamma_b$ spikes that the first residual block receives as input are propagated through both the direct and the residual path. Therefore, the number of spikes at the output of the first residual block is doubled. In consequence, the following layer receives $2\times \gamma_b$ spikes at its input. Similarly, the two flows of spikes add up at the output of the second residual block that generates $4 \times \gamma_b$ spikes. The number of spikes increases exponentially as the network becomes deeper, thus creating an \emph{avalanche} of spikes that have to be processed by the final layers of the network. To overcome this problem we propose a modification of the ResNet architecture for SNNs, that we call Sparse-ResNet shown in Fig. \ref{fig:spiking_resnets} (d). 

\subsection{Sparse-ResNet} \label{sec:sparse_resnet}
\begin{figure}[htbp]
\centerline{\includegraphics[width=.5\textwidth]{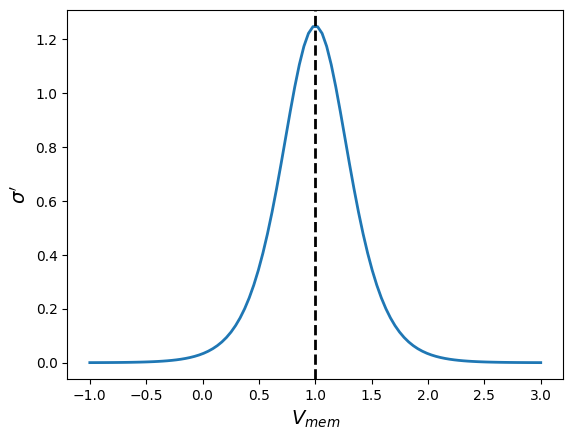}}
\caption{The derivative, $\sigma^\prime$ , of the sigmoid surrogate used in our study: $\sigma^\prime(x, \alpha) = \alpha\,\sigma(\alpha \cdot x)\,(1 - \sigma(\alpha\cdot x))$. Where $\sigma(x)$ is the sigmoid function and $\alpha$ the scaling factor. Here $\alpha=5$ and $V_{th}=1$. The threshold voltage $V_{th}$ is also plotted as a vertical dotted line.}
\label{fig:surrogate}
\end{figure}

Sparse-ResNet makes use of the multi-level spiking neuron both in the direct path like SEW ResNet and after the summation point as Spiking ResNet. The multi-level neuron that is placed after the summation point is called a \emph{barrier neuron}. Its goal is to maintain a low quantization error at the output of the summation point while limiting the number of spikes that are generated. However, as pointed out in \cite{fang_deep_2021}, placing a spiking neuron just after the summation point causes a performance degradation for the directly trained spiking ResNets. The vanishing/exploding gradients, induced by the surrogate function of the spiking neuron, are indeed the main causes of the performance degradation observed in \cite{fang_deep_2021}. 
SEW-ResNets solve this problem by removing the spiking neuron after the summation point. Therefore, since $S_l = O_l$ and the $g$ function is the addition, the gradients that are back-propagated both in the direct, $A_l$ and the residual connections, $R_l$, are equal to the incoming gradient $\frac{\partial L}{\partial O_l}$:
\begin{equation}\label{eq:sew_gradient}
    \frac{\partial L}{\partial A_l} = \frac{\partial L}{\partial R_l} = \frac{\partial L}{\partial O_l}
\end{equation}
In the Sparse-ResNet model instead, the incoming gradient first goes through the surrogate gradient function of the spiking neuron before being propagated through the direct and residual connections:
\begin{equation}\label{eq:sparse_gradient}
\frac{\partial L}{\partial A_l} = \frac{\partial L}{\partial R_l} = \frac{\partial L}{\partial O_l} \, \frac{\partial O_l}{\partial S_l} = \frac{\partial L}{\partial O_l} \, \sigma^\prime (S_l - V_{th})
\end{equation}
Commonly used surrogate functions \cite{neftci_surrogate_2019}, like the one used in our study and shown in Fig. \ref{fig:surrogate}, are highly peaked at $V_{th}$ and drops quickly as the membrane potential of the neuron, $V_{mem}$ moves away from the threshold voltage.

Since the neuron that follows the summation point directly receives the unweighted spikes $S_l(t)$ as input, its membrane potential is either close to zero or higher than $V_{th}$. As it can be seen from Fig. \ref{fig:surrogate}, in both cases the derivative $\sigma^\prime$ has very low amplitude, thus creating the conditions for the gradient to vanish as already observed in \cite{fang_deep_2021}. 
To overcome this vanishing gradient problem we replace the surrogate gradient function of the barrier neurons with a Straight-Through-Estimator (STE) \cite{bengio_estimating_2013}. The STE introduces a far larger deviation from the \emph{true gradients} than a surrogate function like the one shown in Fig. \ref{fig:surrogate}. But, the rational behind our choice is that we can accept some deviation from the true gradients in exchange for a higher gradient amplitude. Moreover, by using the STE the gradient of the barrier neurons becomes:
\begin{equation}
    \frac{\partial O_l}{\partial S_l} = \sigma^\prime(S_l - V_{th}) = 1
\end{equation}
By replacing the previous gradient into Eq. \ref{eq:sparse_gradient} we can observe that the gradients that are back-propagated through the direct and the residual connections are now equal for both SEW-ResNets and Sparse-ResNets. We provide experimental results in Sec. \ref{sec:gradient_flow} and Sec. \ref{sec:propagation_residual} to confirm our analysis on the gradients and the sparsity for both Sparse-ResNets and SEW-ResNets.

\section{Experimental Results}
\subsection{Datasets and Models} \label{sec:dataset_models}
We train the SNN models using SG training method on two different image classification datasets (CIFAR-10, CIFAR-100) \cite{Krizhevsky09learningmultiple} as well as a neuromorphic data classification dataset (CIFAR-10-DVS) \cite{li_cifar10-dvs_2017}. 

The CIFAR-10/100 datasets are made of 32x32 RGB images representing 10 and 100 classes respectively. For CIFAR-10/100 we train the SNNs using stochastic gradient descent (SGD), with a learning rate of $8 \cdot 10^{-2}$. The learning rate is exponentially decayed with a factor of 0.9 each 50 epochs, and data augmentation (random resize and horizontal flip) is used. Each network is trained for 1500 epochs.

The CIFAR-10-DVS neuromorphic dataset is obtained by recording the moving images of the CIFAR-10 dataset on a LCD monitor by a DVS camera. The DVS camera has a spatial resolution of $128 \times 128$. The event-to-frame integration method described in \cite{fang_incorporating_2021} is used for pre-processing CIFAR-10-DVS.
The neuromorphic data are then represented by $T$ frames which contain the summation of the events, in both positive and negative polarities, in each temporal slice. Similar event-to-frame integration methods are widely adopted in SNN research \cite{kaiser_synaptic_2020, lee_training_2016, xing_new_2020}. The dataset is composed of 1K images per class for a total of 10K images. In our setting, 80\% of the images are used for training and the rest for testing the models. We train Sparse-ResNet18 and VGG16 on the CIFAR-10-DVS using the Adam optimizer and a learning rate of $10^{-3}$ that is exponentially decayed with a factor of 0.5 each 50 epochs. Finally, the network is trained for 500 epochs.

\subsection{Network activity and sparsity}
The SNN activity is defined as the total number of spikes generated by the SNN during the inference, that is during $T$ timesteps. Let us call $z_l(t)$, of dimension $(T, C, H, W)$, the output of the $l^{th}$ layer of an SNN with total depth $L$. Then the total activity of the layer $l$ is computed as follows:
\begin{equation}
   \# Spikes_l = \displaystyle\sum^{T} \sum^{C} \sum^{H} \sum^{W} z_l(t)|_{[0,1]}
\end{equation}
Where $z_l(t)|_{[0,1]}$ represents the value of $z(t)$ restricted to the interval $[0, 1]$. 
Finally, the total activity of the SNN is computed by summing the activities of each layer as follows:
\begin{equation}
    \# Spikes_{SNN} = \displaystyle\sum^l \# Spikes_l 
\end{equation} \label{eq:theta}
The activity defined above measures the total number of events, i.e. spikes, that have to be processed and transmitted for each inference. As discussed in the next section, this metric is tightly related to the SNN energy consumption.

\subsection{Energy estimation}
The SNN energy consumption is estimated using the metric introduced in \cite{lemaire_analytical_2023}. 
The amount of energy consumed by the SNN during the inference is computed by taking into account the network architecture, the number of parameters and activations as well as the sparsity. Based on these information, the energy model outputs the energy consumed by the synaptic operations, the memory accesses and addressing. 

Memory accesses are computed based on the assumption that each SNN layer has its own local Static RAM (SRAM) memory. This local memory stores the layer parameters (weights and biases) and also acts as a buffer for the spikes emitted by the spiking neurons. 

Synaptic operations are typically considered to be accumulations (ACC) for SNNs. This is indeed the case when $N=1$ (binary spikes) as the synaptic weight is accumulated at most once each timestep into the membrane potential. In such a case we consider the energy cost of a synaptic operation to be equal to one ACC operation. However when $N > 1$ (multi-level spikes), the synaptic weight could be accumulated up to $N$ times for each received spike. In such a case, we assume the worst case hypothesis described above, and consider that the synaptic energy cost of a multi-level spike corresponds to $N \times \text{ACC}$ operations, whatever the exact value carried by each spike. This overestimation is negligible as we discuss in Sec. \ref{sec:energy-cifar-10} and \ref{sec:energy_neuromorphic}.

\subsection{Comparison to previous works on image classification}
In this section our proposed method is compared with state-of-the-art results on the CIFAR-10/100 image datasets. The experimental results are provided in Tab. \ref{tab:sota-c10}. To provide a comprehensive comparison we include results from different training methods. In ANN/QANN2SNN methods the resulting SNN is obtained by converting either an FP ANN or a quantized ANN respectively. On the other hand, in SG methods the SNN is directly trained in the spiking domain with surrogate gradients. As it can be observed from Tab. \ref{tab:sota-c10}, the number of timesteps, i.e. the latency, can be drastically reduced from ANN2SNN to QANN2SNN and SG methods. These results highlight the importance of taking into account the quantization noise during the training process to achieve both high accuracy and low latency. We can also observe that, for all datasets and network architectures, the adoption of multi-level neurons allows further reducing the latency compared to binary SNNs. Our results indeed show that using only 1 timestep we can obtain the best accuracy results compared to both binary and others ternary or multi-level SNNs with equal or similar network architectures. On CIFAR-100 for instance we improve by more than 3\% the accuracy compared to IM-Loss \cite{guo_im-loss_2022} while reducing the latency by a factor of 5. Similar trends can be observed for the CIFAR-10 dataset where, for example, we improve the accuracy by 0.35\% and reduce the latency by 4, compared to RecDis-SNN \cite{guo_recdis-snn_2022} using ResNet18/19 respectively.

\begin{table*}
\caption{Comparison with SoTA methods on CIFAR-10/100. Ours results are given for $N=4$.}\label{tab:sota-c10}
\begin{tabular*}{\textwidth}{@{} ccccccc@{} }
\toprule
Dataset & Method & Type & Neuron & Architecture & Timestep & Accuracy\\
\midrule
\multirow{16}{*}{\rotatebox[origin=c]{90}{CIFAR-10}} & RMP-SNN \cite{han_rmp-snn_2020} & ANN2SNN & binary & VGG16 & 2048 & 93.63\% \\
& QFFS \cite{li_quantization_2022} & QANN2SNN & binary & VGG16 & 4 & 92.64\% \\
& QCFS \cite{ding_optimal_2021} & QANN2SNN & binary & VGG16 & 4 & 93.86\% \\
& DIET-SNN \cite{rathi_diet-snn_2021} & SG & binary & VGG16 & 10 & 93.44\% \\
& IM-Loss \cite{guo_im-loss_2022} & SG & binary & VGG16 & 5 & 93.85\% \\
& ATIF \cite{castagnetti_trainable_2023} & SG & binary & VGG16 & 4 & 93.13\% \\
& \textbf{Ours} & SG & \textbf{multi-level} & VGG16 & \textbf{1} & \textbf{94.34}\% \\
\cmidrule{2-7}
& S-ResNet \cite{hu_spiking_2023} & ANN2SNN & binary & ResNet20 & 350 & 91.82\% \\
& ACP \cite{li_free_2021} & ANN2SNN & binary & ResNet20 & 4 & 84.70\% \\
& QFFS \cite{li_quantization_2022} & QANN2SNN & binary & ResNet18 & 4 & 93.14\% \\
& DIET-SNN \cite{rathi_diet-snn_2021} & SG & binary & ResNet20 & 10 & 92.54\% \\
& IM-Loss \cite{guo_im-loss_2022} & SG & binary & ResNet19 & 4 & 95.40\% \\
& RecDis-SNN \cite{guo_recdis-snn_2022} & SG & binary & ResNet19 & 4 & 95.53\% \\
& Ternary Spike \cite{guo_ternary_2023} & SG & ternary & ResNet20 & 4 & 94.96\% \\
& MLF \cite{feng_multi-level_2022} & SG & multi-level & ResNet20 & 4 & 94.25\% \\
& MBLIF \cite{xiao_multi-bit_2024} & SG & multi-level & ResNet20 & 1 & 94.59\% \\
& \textbf{Ours} & SG & \textbf{multi-level} & SEW-ResNet18 & \textbf{1} & \textbf{95.94}\% \\
& Ours & SG & multi-level & Sparse-ResNet18 & 1 & 95.69\% \\
\cmidrule{2-7}
\multirow{10}{*}{\rotatebox[origin=c]{90}{CIFAR-100}} & RMP-SNN \cite{han_rmp-snn_2020} & ANN2SNN & binary & VGG16 & 2048 & 70.93\% \\
& QCFS \cite{ding_optimal_2021} & QANN2SNN & binary & VGG16 & 4 & 69.62\% \\
& DIET-SNN \cite{rathi_diet-snn_2021} & SG & binary & VGG16 & 5 & 69.67\% \\
& IM-Loss \cite{guo_im-loss_2022} & SG & binary & VGG16 & 5 & 70.18\% \\
& \textbf{Ours} & SG & \textbf{multi-level} & VGG16 & \textbf{1} & \textbf{73.75}\% \\
\cmidrule{2-7}
& DIET-SNN \cite{rathi_diet-snn_2021} & SG & binary & ResNet20 & 5 & 64.07\% \\
& RecDis-SNN \cite{guo_recdis-snn_2022} & SG & binary & ResNet19 & 4 & 74.10\% \\
& Ternary Spike \cite{guo_ternary_2023} & SG & ternary & ResNet20 & 4 & 74.02\% \\
& MBLIF \cite{xiao_multi-bit_2024} & SG & multi-level & ResNet20 & 1 & 75.43\% \\
& Ours & SG & multi-level & SEW-ResNet18 & 1 & 75.27\% \\
& \textbf{Ours} & SG & \textbf{multi-level} & Sparse-ResNet18 & \textbf{1} & \textbf{75.7}\% \\
\bottomrule
\end{tabular*}
\end{table*}

Moreover, it can be observed that Sparse-ResNet achieves similar or even slightly better performance results compared to SEW-ResNet, thus validating our approach based on a barrier neuron with STE and described in Sec. \ref{sec:sparse_resnet}. More results are provided in Sec. \ref{sec:propagation_residual} to show that Sparse-ResNet also improves the sparsity compared to SEW-ResNet. Moreover, in Sec. \ref{sec:gradient_flow} we provide experimental results that validate the effectiveness of the STE in reducing the vanishing gradient problem in spiking residual networks.

\subsection{Comparison to previous works on Neuromorphic data classification}
The experimental results obtained for the CIFAR-10-DVS dataset are shown in Tab. \ref{tab:sota-c10-dvs}. Here we report the results for a multi-level Sparse-ResNet18 and VGG16 for the $[T=1, N=10]$ configuration which provides 10 quantization intervals. As we can see from Tab. \ref{tab:sota-c10-dvs} we can reach close to the state of the art accuracy while decreasing the latency to only 1 timestep. Previous state of the art methods, both with binary and multi-level neurons, are significantly outperformed as they require a latency 10 to 20 times higher to get a similar performance level.

\begin{table*}
\caption{Comparison with SoTA methods on CIFAR-10-DVS. Ours results are given for $N=10$.}\label{tab:sota-c10-dvs}
\begin{tabular*}{\textwidth}{@{} ccccccc@{} } 
\toprule
Dataset & Method & Type & Neuron & Architecture & Timestep & Accuracy\\
\midrule
\multirow{9}{*}{\rotatebox[origin=c]{90}{CIFAR-10-DVS}}
& DSR \cite{meng_training_2022} & SG & binary & VGG11 & 20 & 77.27\% \\
& Dspike \cite{li_differentiable_2021} & SG & binary & ResNet18 & 10 & 75.4\% \\
& MSG \cite{li_directly_2024} & SG & binary & ResNet18 & 10 & 79.35\% \\
& SEW-ResNet \cite{fang_deep_2021} & SG & binary & Wide-7B-Net & 16 & 74.4\% \\
& IM-Loss \cite{guo_im-loss_2022} & SG & binary & ResNet19 & 10 & 72.6\% \\
& RecDis-SNN \cite{guo_recdis-snn_2022} & SG & binary & ResNet19 & 10 & 72.42\% \\
& MLF \cite{feng_multi-level_2022} & SG & multi-level & ResNet14 & 10 & 70.36\% \\
& Ternary Spike \cite{guo_ternary_2023} & SG & ternary & ResNet20 & 10 & 79.8\% \\
& Ours & SG & multi-level & Sparse-ResNet18 & 1 & 79.1\% \\
& Ours & SG & multi-level & VGG16 & 1 & 76.97\% \\
\bottomrule
\end{tabular*}
\end{table*}

More results on sparsity and energy consumption will be provided in Sec. \ref{sec:energy_neuromorphic}, to assess wether the reduction in latency also translates in energy efficiency improvement.

\section{Ablation studies}
\subsection{Energy efficiency comparison between multi-level and binary spiking neurons on Image classification} \label{sec:energy-cifar-10}
Some experiments were conducted to compare the energy efficiency of binary and multi-level SNNs on the CIFAR-10 dataset. To do so, different configurations of the VGG16 architecture were studied, both in terms of latency $T$ and spikes value $N$. In the first configuration we fix $N=1$ and train the VGG16 network with different number of timesteps in the range $T\in[1, 10]$. In the second configuration, we set the number of timestep to $T=1$ and the value of the spikes in the range $N\in[1, 10]$. This setup allows comparing configurations with equal number of quantization intervals. As an example the configurations $[T=4, N=1]$ and $[T=1, N=4]$ both provide the same number of quantization intervals. However, in the first case the quantization process takes place through time by processing multiple binary spikes. In the second case the discretization process is performed in one timestep through multi-valued spikes. For each configuration, SNNs are trained using the setup described in Sec. \ref{sec:dataset_models}.

\begin{figure}[htbp]
\centerline{\includegraphics[width=.5\textwidth]{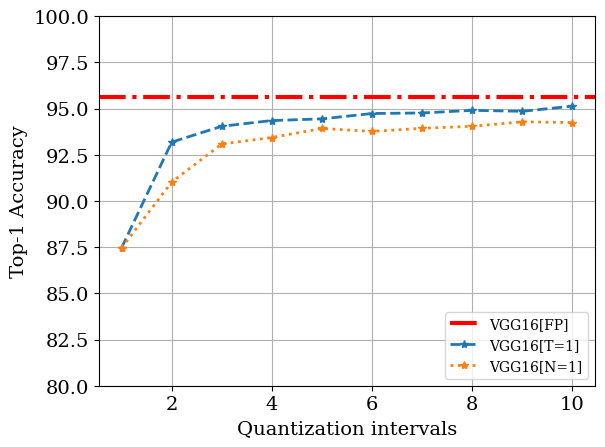}}
\caption{\textbf{Top-1 accuracy of VGG16 on the CIFAR-10 dataset}. The horizontal dotted line represents the accuracy of the ANN FP version of the network.}
\label{fig:cifar_10_accuracy}
\end{figure}

The accuracy of the different SNNs configurations as well as the ANN baseline is shown in Fig. \ref{fig:cifar_10_accuracy}. As expected, the accuracy gap between the ANN and the SNNs reduces when increasing the number of quantization intervals. Moreover, as it can be observed, there is a functional equivalence between both SNN configurations. For a given number of quantization intervals, the accuracy converges to very close values. These results thus confirm the analysis of Sec. \ref{sec:quantization}, where we have shown that the same quantization function can be obtained with different $[N, T]$ configurations. 

\begin{figure*}[htbp]
\centerline{\includegraphics[width=.9\textwidth]{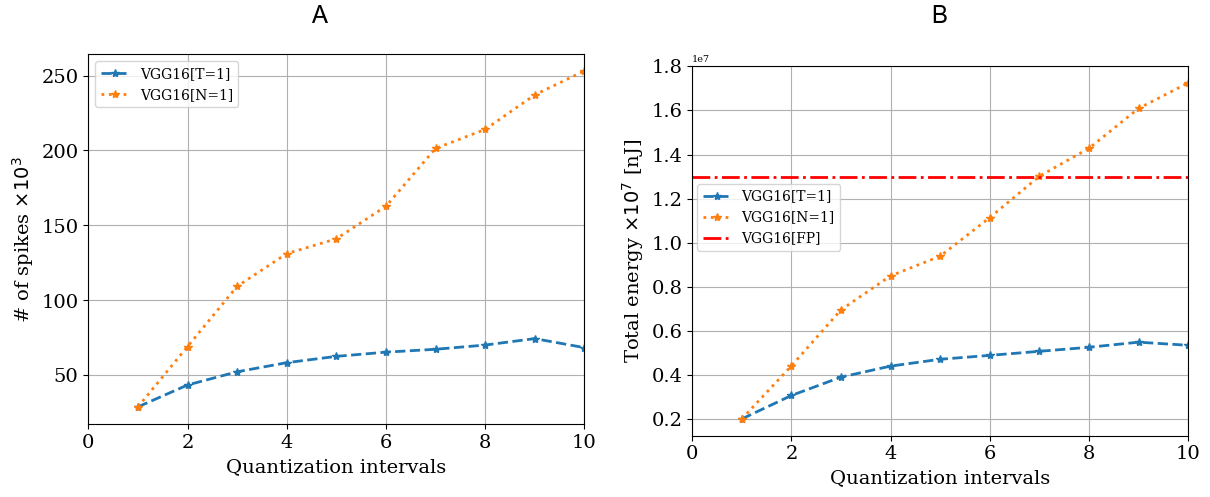}}
\caption{\textbf{Spiking activity and total energy consumption of SNNs and ANN for the CIFAR-10 dataset}. We provide (A) the total spiking activity of VGG16 for different configurations of multi-level SNNs $\{T=1,\,N\in[1, 10]\}$ and binary SNNs $\{T\in[1,10],\,N=1\}$. (B) The total energy consumption of the different SNN configurations. The horizontal dotted line represents the total energy consumption of the ANN.}
\label{fig:cifar_10_spikes_energy}
\end{figure*}

The total spiking activity as well as the total energy consumption are shown in Fig. \ref{fig:cifar_10_spikes_energy} (A) and Fig. \ref{fig:cifar_10_spikes_energy} (B) respectively.
Even though the different binary and multi-level configurations are functionally equivalent, we can observe that they are considerably different both in terms of spiking activity and energy consumption.
As an example, the $[T=4,N=1]$ configuration generates $130\cdot10^3$ binary spikes for each inference, while the same functional configuration with multi-level neurons, $[T=1,N=4]$, only produces $57\cdot10^3$ valued spikes. That is $43\%$ less spikes that have to be processed for each input image. Moreover, it can be observed that for the binary SNNs the total activity grows almost linearly with the timesteps while there is a (sub)logarithmic increase for the multi-level configurations. The spiking activity directly impacts the total energy consumption shown Fig. \ref{fig:cifar_10_spikes_energy} (B). It can be observed that even if the binary SNNs are more energy efficient than the ANN at low-timesteps, e.g. the binary SNNs provides 35\% energy reduction when $[T=4,N=1]$, the gap quickly reduces as T increases. At $T=8$ for instance, the binary SNNs already consumes $10\%$ more energy than the ANN. On the other hand, the multi-level VGG16 maintains a high level of energy efficiency at each operating point. As an example, when $[T=1,N=4]$ there is a 66\% energy reduction compared to the ANN. At the $[T=1,N=8]$ operating point the energy reduction is still 60\% compared to the ANN.

\begin{table*}
\caption{Energy consumption breakdown for binary and multi-level VGG16 SNNs and ANN on CIFAR-10. All energy values are expressed in $[nJ]$. $E_N$ corresponds to the total energy consumption of the multi-level SNN, while $E_{binary}$ refers to the binary SNN. The symbol $[\downarrow]$ means lower is better.}\label{tab:energy-c10}
    
\begin{tabular*}{\textwidth}{@{} cccccc@{} } 
\toprule
&  \multicolumn{1}{c}{\textbf{ANN}} & \multicolumn{4}{c}{\textbf{SNN}}\\
\midrule
& & binary & multi-level & binary & multi-level \\
& & $[T=4,N=1]$ & $[T=1,N=4]$ & $[T=8,N=1]$ & $[T=1,N=8]$ \\
\textbf{Memory} & & & & &\\
Potentials & - & $6.0\cdot10^6$ & $3.1\cdot10^6$ & $10.0\cdot10^6$ & $3.6\cdot10^6$ \\
Weights & $5.58\cdot10^6$ & $2.4\cdot10^6$ & $1.3\cdot10^6$ & $4.1\cdot10^6$ & $1.54\cdot10^6$ \\
Bias & $6.9\cdot10^2$ & $27.8\cdot10^2$ & $6.9\cdot10^2$ & $55.7\cdot10^2$ & $6.9\cdot10^2$\\
In/Out & $6.1\cdot10^6$ & $3.5\cdot10^3$ & $1.54\cdot10^3$ & $5.7\cdot10^3$ & $1.8\cdot10^3$ \\
Total & $11.6\cdot10^6$ & $8.4\cdot10^6$ & $4.38\cdot10^6$ & $14.2\cdot10^6$ & $5.19\cdot10^6$ \\
\hline
\makecell{\textbf{Synaptic} \\ \textbf{Operations}} & $1.29\cdot10^6$ & $13.3\cdot10^3$ & $27.9\cdot10^3$ & $22.4\cdot10^3$ & $65.4\cdot10^3$ \\
\hline
\textbf{Addressing} & $4.9\cdot10^2$ & $9.0\cdot10^2$ & $5.5\cdot10^2$ & $14.8\cdot10^2$ & $8.7\cdot10^2$ \\
\hline
\textbf{Total} & $\mathbf{12.9\cdot10^6}$ & $\mathbf{8.5\cdot10^6}$ & $\mathbf{4.41\cdot10^6}$ & $\mathbf{14.28\cdot10^6}$ & $\mathbf{5.26\cdot10^6}$ \\
\hline
$\mathbf{E_{N}/E_{binary}}$ $[\downarrow]$ & & \multicolumn{2}{c}{0.51} & \multicolumn{2}{c}{0.37}\\
\hline
$\mathbf{E_{SNN}/E_{ANN}}$ $[\downarrow]$ & & 0.65 & 0.34 & 1.10 & 0.4\\
\bottomrule
\end{tabular*}
\end{table*}

Finally, the complete energy consumption breakdown is shown in Tab. \ref{tab:energy-c10}. Here we provide the detailed estimation for the ANN and the configurations with 4 and 8 quantization intervals for both the binary and the multi-level SNNs. As it can be seen, the total energy consumption, especially for the SNNs, is strongly dominated by the cost of memory accesses. Each spike generated by a neuron leads to three memory accesses (two read operations for retrieving the weights and the current value of the membrane potential and a write operation) and a synaptic operation (ACC or $N \times \text{ACC}$ for the binary and multi-valued case respectively). The energy cost for one memory access is on average 10 to 100 times higher than a synaptic operation \cite{jouppi_ten_2021}. Indeed, we can observe differences of two to three orders of magnitude between the energy consumed by the memory accesses and the synaptic operations. We can also observe that, as expected, the energy consumed by the synaptic operations is greater for the multi-valued spike. However, as the amount of generated spikes is lower than binary spikes, this increase is largely counterbalanced by the decrease of energy required for accessing the memory. These results clearly confirm that the total amount of spikes, shown in Fig. \ref{fig:cifar_10_spikes_energy} (1) is the most important metric to consider for reducing the energy consumption of SNNs. Neither the average activity per timestep nor the latency alone are sufficient to properly evaluate the overall energy consumption. In that sense our results show that, using multi-valued spikes, the energy efficiency of SNNs can be greatly improved by 2 to 3 times depending on the number of quantization intervals.

\subsection{Energy efficiency comparison between multi-level and binary spiking neurons on Neuromorphic data classification} \label{sec:energy_neuromorphic}
In this section we study two configurations of the VGG16 architecture, with multi-level and binary neurons, on the CIFAR-10-DVS neuromorphic classification dataset. Specifically we compare the multi-level VGG16 network given in Tab. \ref{tab:sota-c10-dvs} with a binary VGG16 trained on $T=10$ timesteps. Both configurations thus provide 10 quantization intervals as well as similar accuracy levels: 76.97\% when $\{T=1,\,N=10\}$ and 75.5\% for the binary configuration $\{T=10,\,N=1\}$.
The energy breakdown for both SNNs as well as the ANN is given in Tab. \ref{tab:energy-c10-dvs}.

\begin{table*}
\caption{Energy consumption breakdown for binary and multi-level VGG16 SNNs and ANN on CIFAR-10-DVS. All energy values are expressed in $[nJ]$. $E_N$ corresponds to the total energy consumption of the multi-level SNN, while $E_{binary}$ refers to the binary SNN. The symbol $[\downarrow]$ means lower is better.}\label{tab:energy-c10-dvs}

\begin{adjustbox}{center}
\begin{tabular*}{.65\textwidth}{@{} cccc@{} } 
\toprule
&  \multicolumn{1}{c}{\textbf{ANN}} & \multicolumn{2}{c}{\textbf{SNN}}\\
\midrule
& & binary & multi-level \\
& & $[T=10,N=1]$ & $[T=1,N=10]$ \\
\textbf{Memory} & & & \\

Potentials & - & $233.3\cdot10^6$ & $90.6\cdot10^6$ \\
Weights & $52\cdot10^6$ & $32.2\cdot10^6$ & $12.8\cdot10^6$ \\
Bias & $6.9\cdot10^2$ & $69.7\cdot10^2$ & $6.9\cdot10^2$ \\
In/Out & $186.3\cdot10^6$ & $40.5\cdot10^3$ & $15.8\cdot10^3$ \\
Total & $238.3\cdot10^6$ & $265.5\cdot10^6$ & $103.4\cdot10^6$ \\
\hline
\makecell{\textbf{Synaptic} \\ \textbf{Operations}} & $12.1\cdot10^6$ & $170.8\cdot10^3$ & $664.6\cdot10^3$ \\
\hline
\textbf{Addressing} & $6.2\cdot10^3$ & $10.3\cdot10^3$ & $7.7\cdot10^3$ \\
\hline
\textbf{Total} & $\mathbf{250.4\cdot10^6}$ & $\mathbf{265.7\cdot10^6}$ & $\mathbf{104.1\cdot10^6}$ \\
\hline
$\mathbf{E_{N}/E_{binary}}$ $[\downarrow]$ & & \multicolumn{2}{c}{0.39} \\
\hline
$\mathbf{E_{SNN}/E_{ANN}}$ $[\downarrow]$ & & 1.06 & 0.41 \\
\bottomrule
\end{tabular*}
\end{adjustbox}
\end{table*}

First it can be observed that the total energy consumption for both the SNNs and the ANN is almost twenty time higher than the CIFAR-10 results shown in Tab. \ref{tab:energy-c10}. The higher resolution of CIFAR-10-DVS ($128 \times \ 128$) compared to CIFAR-10 ($32 \times \ 32$) leads to an increase of the number of memory accesses and operations to process the input frames as well as the feature maps of hidden layers. However, results shown in Tab. \ref{tab:energy-c10-dvs} confirm the energy efficiency improvements of multi-level SNNs, even on neuromorphic data. As it can be observed, the multi-level VGG16 reduces by 60\% the energy consumption compared to the ANN. The energy gains come from the reduction of energy related to memory transfers. The multi-level SNN consumes in total $103.4\cdot10^6$ nJ for the memory accesses, while $238.3\cdot10^6$ nJ are required for the ANN. In the multi-level SNN, most of the energy is consumed for accessing the neuron potentials ($90.6\cdot10^6$ nJ) while only $12.8\cdot10^6$ nJ are related to the memory accesses associated with the synaptic weights. On the other hand, the ANN consumes four times more energy for accessing the synaptic weights ($52\cdot10^6$ nJ) and twice more for accessing the feature maps, i.e. In/Out $186.3\cdot10^6$ nJ. Finally, there is a difference of almost three order of magnitude in the energy consumption of the synaptic operations between the ANN and the SNNs. However, for the ANN the energy related to the synaptic operations accounts for less than 5\% of the total energy budget. Our results are close to those already observed in \cite{dampfhoffer_are_2023}, that is the energy cost of synaptic operations is very small compared to the one of transferring data from the memory. Therefore, the energy gains obtained by reducing the cost of the synaptic operations are marginal when considering the total energy consumption. Finally, the binary SNN does not exhibit energy gains compared to the ANN when using latencies that provide state of the art accuracy results on the CIFAR-10-DVS dataset, i.e. 10 or more timesteps. When $T=10$ the SNN fires $1.5\cdot10^6$ spikes while only $591\cdot10^3$ spikes are emitted when $N=1$. 

As shown in Tab. \ref{tab:energy-c10-dvs}, it represents a 60\% reduction in the total spiking activity which translates in an equivalent energy gain between multi-level and binary SNNs. These results again confirm that, even for neuromorphic datasets, the energy consumption can be lowered by reducing the total amount of spikes generated by the SNN.

\subsection{Gradient propagation in spiking ResNet architectures} \label{sec:gradient_flow}
\begin{figure}[htbp]
\centerline{\includegraphics[width=.6\textwidth]{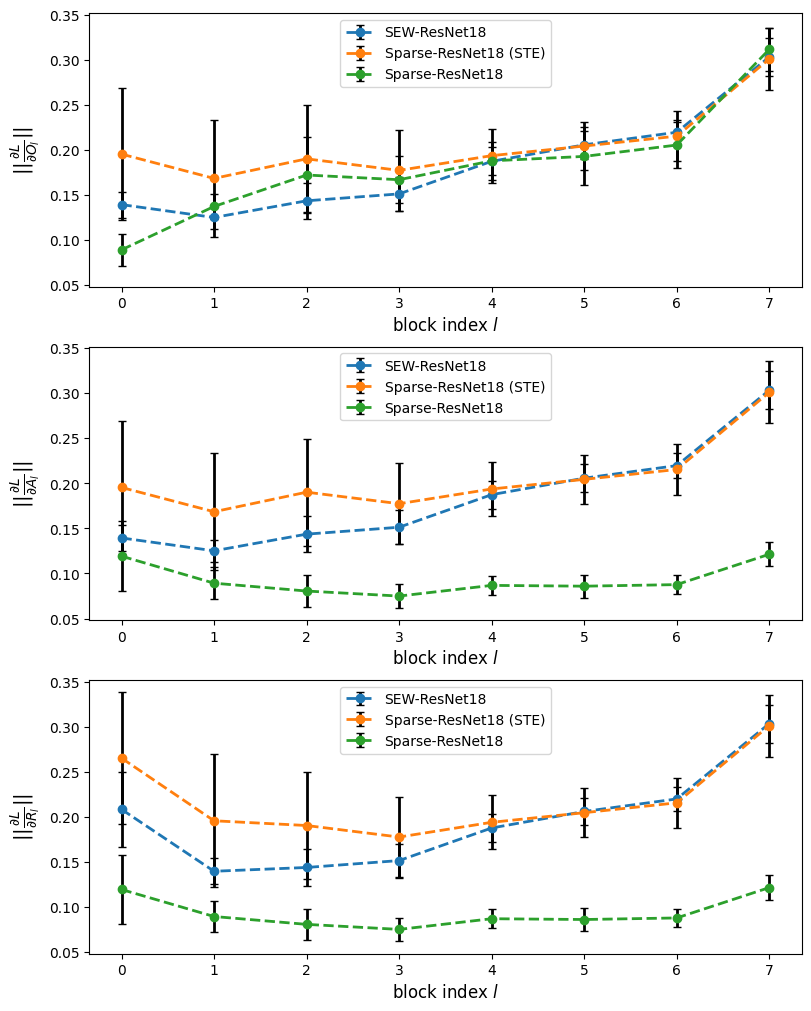}}
\caption{\textbf{Using STE helps overcome the vanishing gradient}. The gradients of the residual blocks in SEW-ResNet18, Sparse-ResNet18 with STE and Sparse-ResNet18 without STE on the CIFAR-10 dataset. The block 0 is the first block after the input.}
\label{fig:resnet_gradient}
\end{figure}

\begin{figure}[htbp]
\centerline{\includegraphics[width=.5\textwidth]{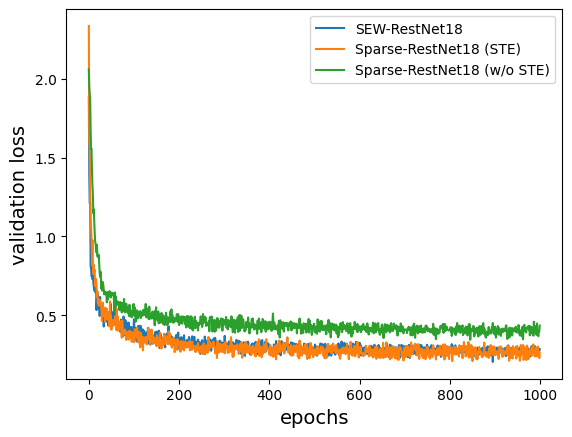}}
\caption{Comparison of the validation loss of SEW-REsNet18, Sparse-ResNet18 with STE and Sparse-ResNet18 without STE on CIFAR-10.}
\label{fig:resnet_loss}
\end{figure}

In this section we study the gradient propagation in the spiking residual architectures SEW-ResNet and the Sparse-ResNet that has been introduced in Sec. \ref{sec:sparse_resnet}. To assess the improvement provided by the STE we also consider a variant of Sparse-ResNet where all the spiking neurons, including the barrier neurons, use the same surrogate derivative functions shown in Fig. \ref{fig:surrogate}. For each block of the three architectures, SEW-REsNet18, Sparse-ResNet18 with STE and Sparse-ResNet18 without STE we measure the gradient norm on the direct path, $\frac{\partial L}{\partial A_l}$, the residual path $\frac{\partial L}{\partial R_l}$ and after the summation point $\frac{\partial L}{\partial O_l}$. The gradient is averaged over $10^4$ minibatches of the CIFAR-10 dataset. We also repeat the simulation by training the networks 10 times starting from different initial conditions. The gradient mean and standard deviation shown in Fig. \ref{fig:resnet_gradient} are provided for each residual block of the network .
The experimental results confirm the analysis given in Sec. \ref{sec:sparse_resnet}. As it can be observed, the gradients norm on the direct and residual paths of SEW-ResNet and Sparse-ResNet with STE are closer to each other and consistently higher than the gradients of Sparse-ResNet without STE. Indeed, when STE is not used the updates of the gradient on Sparse-ResNet are weaker, thus slowing down the training and decreasing the performance. This behavior can be observed in Fig. \ref{fig:resnet_loss} where we show the validation loss, computed at the end of each training epoch for the three networks on the CIFAR-10 dataset. 
The loss of Sparse-ResNet without STE declines more slowly and finally reaches a higher value compared to SEW-ResNet and Sparse-ResNet with STE.

\subsection{Spikes propagation in residual layers} \label{sec:propagation_residual}
\begin{figure}[htbp]
\centerline{\includegraphics[width=.5\textwidth]{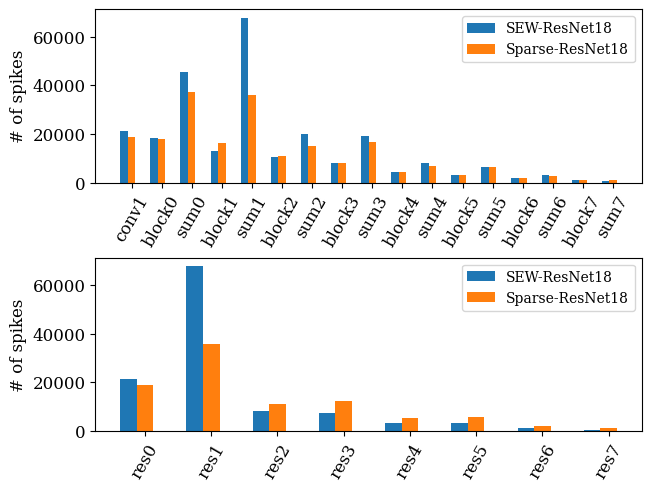}}
\caption{\textbf{Layer-wise activity of SEW-ResNet18 and Sparse-ResNet18}. Top: number of spikes generated in the direct path ($block_i$) and after the summation point ($sum_i$). Bottom: number of spikes generated in the residual path ($res_i$). The values are averaged over all the images of the CIFAR-10 test set.}
\label{fig:resnet_sparsity}
\end{figure}

In this section we provide some experimental results to assess the effectiveness of the barrier neurons in limiting the spike propagation in the residual layers. To do so, we compare the activity of SEW-ResNet and Sparse-ResNet with STE measured on the CIFAR-10 dataset. Both networks are trained using the $[T=1, N=4]$ configuration, that provides almost the same accuracy, as shown in Tab. \ref{tab:sota-c10}. We measure the number of generated spikes in the direct and the residual paths for both networks. The spikes that are generated at the summation point are computed as the sum of the number of spikes coming from the direct path and the spikes coming from the residual path for SEW-ResNet. For Sparse-ResNet, we measure instead the number of spikes after the barrier neuron. Our measure corresponds, in both cases, to the amount of spikes that are transferred from one layer to the next in the context of an event-based communication/processing scheme.
The experimental results are shown in Fig. \ref{fig:resnet_sparsity}, where we can observe that SEW-ResNet consistently generates more spikes in the first layers of the network compared to Sparse-ResNet. The the number of generated spikes stabilizes from the middle to the last layers, where both networks generate almost the same activity. The spike avalanche effect, that we qualitatively described in Sec. \ref{sec:spikes_propagation_qualitative}, can be clearly observed in Fig. \ref{fig:resnet_sparsity}. The number of spikes generated by SEW-ResNet at the first ($sum_0$) and especially at the second ($sum_1$) summation points are considerably higher compared to Sparse-ResNet. 

SEW-ResNet produces 45665 spikes at the summation point $sum_0$ while only 37150 spikes are fired by Sparse-ResNet. The avalanche effect is even more visible at $sum_1$ where there are 67848 spikes for SEW-ResNet and only 35819 for Sparse-ResNet. That is 47\% less activity in one of the layer that generates most of the spikes in the whole network. These results confirm that the barrier neurons of Sparse-ResNet are able to prevent the spike avalanche effect, thus reducing the network activity as we discuss in more details in the next section.

\subsection{Analysis of sparsity in spiking ResNet architectures} \label{sec:resnet_sparsity}
\begin{figure*}[htbp]
\centerline{\includegraphics[width=.9\textwidth]{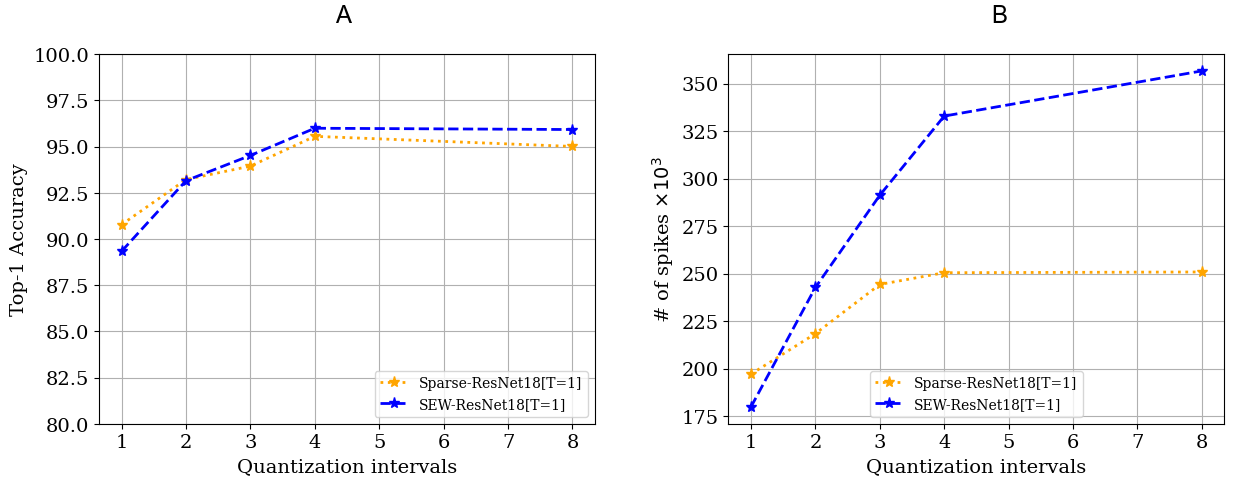}}
\caption{\textbf{(A) Top-1 accuracy and (B) spiking activity of Sparse-ResNet18 and SEW-ResNet18 on the CIFAR-10 dataset.} For all the configurations the number of timesteps $T=1$. The spiking activity is computed by averaging over all the images of the CIFAR-10 test set.}
\label{fig:resnet_accuracy_sparsity}
\end{figure*}

In this final section we provide experimental results on the spiking activity for multi-level SEW-ResNet and Sparse-ResNet. The accuracy as well as the total spiking activity for different configurations on CIFAR-10, i.e. $\{T=1,\,N\in[1, 8]\}$, are shown in Fig. \ref{fig:resnet_accuracy_sparsity}. 

As it can be observed, both networks provide very similar accuracy results, with a marginally better accuracy for SEW-ResNet when $N > 2$. 
However, we can observe that SEW-ResNet generates a much higher level of spiking activity for all the configurations except when $N=1$. As an example, when $N=2$ SEW-ResNet fires 242662 spikes/image while Sparse-ResNet generates 218343 spikes/image, which corresponds to a 10\% reduction in the overall activity while providing very similar accuracy. The spiking activity gap between both network architectures widens as N increases. For the configuration $N=4$, Sparse-ResNet generates 25\% less spikes/image compared to SEW-ResNet with only 0.25\% accuracy drop. The spiking activity reduction provided by Sparse-ResNet rises to 30\% when $N=8$ with an accuracy gap less than 1\%. As discussed in Sec. \ref{sec:spikes_propagation_qualitative} and shown in Fig. \ref{fig:resnet_sparsity}, the difference in spiking activity is mainly due to the spike avalanche effect caused by the first two shortcut connections in the ResNet18 network. This effect would be exacerbated in deeper networks, that is when more consecutive layers use skip connections such as in the ResNet34 architecture \cite{he_deep_2016}, when up to five consecutive residual blocks use shortcut connections. Another example is the MobileNetV2 architectures \cite{sandler_mobilenetv2_2019}, which is composed of 17 consecutive layers of inverted residual blocks. A shortcut connects the input with the output of each block. For the spiking version of that particular architecture, the spike avalanche effect would be significantly higher if the spikes are aggregated at the end of each residual connections as shown in Fig. \ref{fig:avalanche_resnets}. In that case the use of barrier neurons, as proposed in this paper, could lead to a significant reduction in the total spiking activity and thus an energy efficiency improvement.

\section{Conclusion and Future works}
In this paper, we proposed to improve the sparsity and energy efficiency of SNNs both at neuronal and at network level. By leveraging multi-level spiking neurons we have shown that the energy efficiency of SNNs cab be improved. By reducing the quantization noise we are able to provide state of the art results using only 1 timestep, therefore at the lowest possible latency. Our approach can be also applied to neuromorphic data, for which state of the art accuracy were obtained while reducing the latency by a factor of 10. At the architectural level, we identified the spike avalanche effect that impacts most of spiking residual architectures. Sparse-ResNet has been proposed as a solution to the avalanche effect. This residual architecture is indeed able to provide state of the art accuracy results on image classification while reducing by more than 20\% the network activity. We identified several deep residual architectures, widely used in image processing, that can be impacted by the spike avalanche effect. We believe that it would be interesting to extend our approach to these architectures and evaluate the energy efficiency gains, both with realistic energy models and using real neuromorphic hardware.

\bibliographystyle{unsrtnat}
\bibliography{main}

\end{document}